\pdfoutput=1

\documentclass[11pt]{article}

\usepackage{emnlp2021}

\usepackage{times}
\usepackage{latexsym}
\usepackage{multirow}
\usepackage{tikz}
\usepackage{caption}
\usepackage{subcaption}
\usepackage{todonotes}
\usepackage{xcolor}
\usepackage{makecell}
\usepackage{booktabs}
\usepackage[T1]{fontenc}

\usepackage[utf8]{inputenc}

\usepackage{microtype}

\usepackage{color, colortbl}
\definecolor{Gray}{gray}{0.2}

\title{Rethinking Data Augmentation for Low-Resource\\ Neural Machine Translation: A Multi-Task Learning Approach}

\author{Víctor M. Sánchez-Cartagena, Miquel Esplà-Gomis\\
 {\bf Juan Antonio Pérez-Ortiz}, {\bf Felipe Sánchez-Martínez} \\
Dep. de Llenguatges i Sistemes Informàtics \\
Universitat d'Alacant\\
E-03690 Sant Vicent del Raspeig (Spain)\\
{\tt \{vmsanchez,mespla,japerez,fsanchez\}@dlsi.ua.es}}

\date{}

\begin{document}
\maketitle
\begin{abstract}

In the context of neural machine translation, data augmentation (DA) techniques may be used for generating additional training samples when the available parallel data are scarce. Many DA approaches aim at expanding the support of the empirical data distribution by generating new sentence pairs that contain infrequent words, thus making it closer to the true data distribution of parallel sentences. In this paper, we propose to follow a completely different approach and present a multi-task DA approach in which we generate new sentence pairs with transformations, such as reversing the order of the target sentence, which produce unfluent target sentences. During training, these augmented sentences are used as auxiliary tasks in a multi-task framework with the aim of providing new contexts where the target prefix is not informative enough to predict the next word. This strengthens the encoder and forces the decoder to pay more attention to the source representations of the encoder. Experiments carried out on six low-resource translation tasks show consistent improvements over the baseline and over DA methods aiming at extending the support of the empirical data distribution. The systems trained with our approach rely more on the source tokens, are more robust against domain shift and suffer less hallucinations.
\end{abstract}

\section{Introduction}\label{sec:intro}

In order to train reliable neural machine translation (NMT) systems, large amounts of parallel sentences ---sentence pairs in two languages that are mutual translations--- are needed, which constitutes a critical barrier for low-resource language pairs. This problem has been addressed through different approaches, such as transfer-learning from high-resource language pairs~\citep{kocmi-bojar-2018-trivial}, using linguistic annotations~\citep{sennrich-haddow-2016-linguistic}, training multilingual systems~\citep{doi:10.1162} and applying data augmentation strategies~\citep{li-etal-2019-understanding-data,survey2021}, i.e., artificially generating additional parallel sentences.

Data augmentation (DA) is formalised by many authors as a solution to a data distribution mismatch problem~\citep{wang-etal-2018-switchout,wei-etal-2020-uncertainty}. The data distribution of the sentence pairs observed in the training corpus, $\hat{p}(\mathbf{x}, \mathbf{y})$, differs from the true data distribution, $p(\mathbf{x}, \mathbf{y})$. Hence, the system should be trained on a training set that follows $q(\mathbf{x}, \mathbf{y})$, an augmented version of $\hat{p}(\mathbf{x}, \mathbf{y})$ with a wider support. 
In this way,  the trained system is less likely to face totally out-of-distribution data when translating. 

In this paper, we propose a completely different framework for DA in which we generate additional parallel sentences which, despite being completely unlikely under the data distribution, systematically improve the quality of the resulting NMT system. Inspired by one-to-many multilingual NMT, where richer encoder representations are obtained~\citep{dong-etal-2015-multi}, 
we propose a set of simple DA strategies to produce synthetic target sentences aimed at strengthening the encoder. These strategies expose the network during training to new situations where the target-language context is not sufficient to achieve a low loss, and the burden is passed on to the encoder. Recent findings by~\citet{voita2020analyzing} further motivate our approach: they claim that the influence of source tokens in the output predictions of an NMT system decreases as decoding advances. Moreover, to avoid harmful interferences by the out-of-distribution target data generated, we follow a simple multi-task learning (MTL) approach that does not require changes to the model architecture. We call the proposed framework \textbf{multi-task learning data augmentation} (MTL~DA) to stress the fact that the augmented data, which do not follow the distribution of parallel sentences in the training corpus, constitute different auxiliary tasks that nevertheless produce a positive transfer to the main task.

Our framework does not require elaborated preprocessing steps, training additional systems, or data besides the available training parallel corpora. Experiments with six low-resource translation tasks show that it systematically outperforms two powerful methods aiming at extending the support of the empirical data distribution \textcolor{black}{---}RAML~\citep{norouzi2016reward} and  SwitchOut~\citep{wang-etal-2018-switchout}\textcolor{black}{--- and that it can be combined with synthetic corpora generated through back-translation \citep{sennrich2016edinburgh} to get further improvements.}

In the context of explainable deep learning models, we perform an analysis of the relevance of the encoder and decoder representations in the NMT system output, which shows that, thanks to the auxiliary tasks, MTL~DA increases the contribution of the source tokens to the decisions made by the NMT system. Moreover, systems trained with MTL~DA are much more robust against domain shift and produce less hallucinations~\citep{wang-sennrich-2020-exposure}. 

The remainder of the paper is organised as follows. Next section briefly describes our MTL~DA approach and the different auxiliary tasks we evaluated. After that, Sec.~\ref{sec:settings} describes the experimental settings, whereas Sec.~\ref{sec:results} provides and discusses the results obtained. Sec.~\ref{sec:explain} then presents an analysis of the changes in the transformer dynamics induced by our auxiliary tasks as a way of explaining the improvement in translation quality. The paper ends with a review of the most relevant works in the area of DA for NMT in Sec.~\ref{sec:related}, followed by some concluding remarks in Sec.~\ref{sec:conclusions}.

\section{Multi-task learning approach and auxiliary tasks}\label{sec:da}
\label{sec:methods}
We propose a simple MTL approach that consists of using a vanilla NMT system ---in our experiments, it is a transformer system as defined by \citet{vaswani2017attention}--- where all (main and auxiliary) tasks share the encoder and the decoder. In order to avoid harmful interferences by the out-of-distribution target data generated for the auxiliary tasks, we add a task-specific artificial token to the source sentence to constrain the kind of output to be produced~\citep{sennrich2016token,doi:10.1162}, much like in multilingual NMT.
For each auxiliary task, we append a synthetic corpus of the same size to the original training data, which is obtained by applying a transformation to each original pair of sentences. In almost all the tasks, the source sentence is left unchanged while the target sentence is substantially modified.

What follows is a brief explanation of the transformations we have tested and their expected effect on the training dynamics of the encoder. Some of them have been previously applied in DA, but never in an MTL set-up such as the one we are presenting.
Some transformations are controlled by a hyperparameter $\alpha$ that determines the proportion of target words affected by the transformation. In what follows, $t$ denotes the amount of words in the original target sentence. Table~\ref{tab:auxtasks} provides an example of the effect of the different transformations on a single sentence pair.

\begin{table*}[tb]
\centering
\fontsize{10.2}{12.0}\selectfont
\begin{tabular}{lll}
\textbf{Task} & \textbf{Lang.} & \textbf{Synthetic training sample} \\
\hline \hline
\rowcolor{gray!10}
 & source & Es gibt andere Möglichkeiten , die Pyramide zu durchbrechen . \\
 \rowcolor{gray!10}
                            & & \\
                            \rowcolor{gray!10}
                            & & \multirow{-2}{*}{
                           \begin{tikzpicture}
                               \draw (0,0) -- (1,-1);
                               \draw (0.7,0) -- (0.4,-1);
                               \draw (1.5,0) -- (1.6,-1);
                               \draw (3.2,0) -- (2.4,-1);
                               \draw (4.7,0) -- (4.8,-1);
                               \draw (5.8,0) -- (5.7,-1);
                               \draw (6.5,0) -- (2.4,-1);
                               \draw (7.7,0) -- (3.3,-1);
                               \draw (8.9,0) -- (6.0,-1);
                            \end{tikzpicture} }\\
\rowcolor{gray!10}
\multirow{-4}{*}{\makecell[l]{original\\ training\\ sample}}  & target & There 's other ways of breaking the pyramid . \\ \midrule
swap & target & There \textcolor{red}{.} other ways of breaking \textcolor{blue}{pyramid} \textcolor{red}{'s} \textcolor{blue}{the} \\ \hline
token & target & There 's other \textcolor{red}{UNK} of \textcolor{red}{UNK UNK UNK } . \\ \hline
source & target & \textcolor{red}{Es gibt andere Möglichkeiten , die Pyramide zu durchbrechen .} \\ \hline
reverse & target & \textcolor{red}{. pyramid the breaking of ways other 's There} \\ \hline
mono & target & \textcolor{red}{'s There} other ways \textcolor{red}{the pyramid of breaking} . \\ \hline
\multirow{2}{*}{replace} & source & Es gibt \textcolor{red}{aufzurüsten} \textcolor{blue}{kalt} , \textcolor{violet}{Schach} \textcolor{orange}{Spezialwissen} zu durchbrechen . \\ 
                      &  target & There 's \textcolor{red}{arming} \textcolor{blue}{cold} of breaking \textcolor{violet}{chess} \textcolor{orange}{specialties} . \\ 

\hline
\end{tabular}
\caption{A German--English, word-aligned training sample (first row) 
and the result of applying the transformations described in Sec.~\ref{sec:methods} using $\alpha=0.5$ for those transformations controlled by this hyperparameter. 
Words modified by each transformation are coloured; for \emph{swap} and \emph{replace}, a different colour identifies each pair of words that are either swapped or replaced together, respectively.}
\label{tab:auxtasks}
\end{table*}

\makeatletter
\newenvironment{mydescription}%
               {\list{}{\itemsep=1.8pt \leftmargin=7pt
                        \labelwidth\z@ \itemindent-\leftmargin
                        \let\makelabel\descriptionlabel}}%
               {\endlist}
\makeatother

\begin{mydescription}
\item[swap:] Pairs of random target words are swapped until only $(1-\alpha)\cdot t $ words remain in their original position. This task~\citep{artetxe2017unsupervised,lample2018unsupervised}  tries to force the system to trust less the target prefix when generating a new word.


\item[token:] $\alpha \cdot t$ random target words are replaced by a special (UNK)
token~\citep{xie2017data}. 
Again, when generating a new word, the target prefix should become less informative and force 
the system to pay more attention to 
the encoder. This is the effect envisaged by word dropout when preventing posterior collapse in variational autoencoders~\citep{bowman-etal-2016-generating}.  


\item[source:] The target sentence becomes a copy of the source sentence. In this way, the most efficient way of emitting the right output is checking the encoder representation to copy from the source. Some authors have identified such training instances as harmful for NMT~\citep{pmlr-v80-ott18a,khayrallah-koehn-2018-impact}, and only copying in the inverse direction has been proved to be useful~\citep{currey-etal-2017-copied}. However, the MTL framework may allow us to leverage such synthetic training data.


\item[reverse:] The order of the words in the target sentence is reversed. \citet{voita2020analyzing}
suggest that the influence of the encoder decreases along the target sentence; therefore, by reversing the order we expect the system to learn to use more information from the encoder when generating words that usually appear near the end of the sentence. 

\item[mono:] Target words are reordered so as to make the alignment between source and target words monotonous. This transformation 
uses one-to-many word alignments and is inspired by the concept of \emph{biwords} introduced by \citet{sanchez-martinez12a} for the compression of parallel corpora. By making the alignment between source and target words monotonous, the target sentences become less fluent, so we expect the system to pay more attention to the encoder.

\item[replace:] $\alpha \cdot t$ source--target aligned words are selected at random and replaced by random entries in a bilingual lexicon obtained from the training corpus; to this end, one-to-one word alignments are used.\footnote{If the number of aligned words is below $\alpha \cdot t$, all available alignments are used.} This transformation is likely to introduce words that are difficult to produce by relying only on the target language prefix, thus forcing the system to pay attention to the source words. \citet{Fadaee_2017} followed a similar approach; however, they constrained the replacements to produce only fluent target sentences. 
\end{mydescription}

\section{Experimental settings}
\label{sec:settings}

We have conducted experiments for the translation from English to German, Hebrew and Vietnamese, and for the translation in the reverse direction, using corpora commonly used for evaluating DA techniques in low-resource scenarios. We evaluated 
the effect of using each of the MTL~DA auxiliary tasks,
as well as the combination of the best performing ones. We also evaluated two strong DA methods that aim at extending the support of the empirical data distribution by replacing some words by random samples from the vocabulary: SwitchOut (applied on the source side), RAML (applied on the target side), and the combination of both.

\paragraph{Datasets.} Following~\citet{gao2019soft} and \citet{guo-etal-2020-sequence}, for English--German and English--Hebrew we used the training data (speeches of TED and TEDx talks) of the IWSLT 2014 text translation track~\citep{iwslt2014};\footnote{\url{https://sites.google.com/site/iwsltevaluation2014/data-provided}} for development and testing we used the \emph{tst2013} and \emph{tst2014} datasets, respectively. Like~\citet{wang-etal-2018-switchout}, for English--Vietnamese we used the training data (also TED talks) of the IWSLT 2015 text translation track~\citep{iwslt2015};\footnote{\url{https://wit3.fbk.eu/2015-01}} datasets \emph{tst2012} and \emph{tst2013} were used, respectively, for development and testing. 

\textcolor{black}{To evaluate the impact of MTL~DA when it is combined with synthetic data obtained through back-translation \citep{sennrich2016edinburgh} ---that is, by translating a target-language monolingual corpus into the source language---, 
%
we collected additional English monolingual data, back-translated it into German, Hebrew and Vietnamese, and added the resulting synthetic corpus to the training data listed above. 
The monolingual data used consists of all the available monolingual English sentences in the IWSLT 2018 shared task on low-resource MT of TED talks that were not present in the parallel training data described above.
Table~\ref{tab:corpus} provides the amount of sentences and tokens in the training corpora used in our experiments.}

\begin{table}[tb]
\centering
\begin{tabular}{lrrr}

\textbf{Pair} & \textbf{\# sent.} & \textbf{\# left tok.} & \textbf{\#  right tok.} \\

\hline 
\multicolumn{4}{c}{IWSLT parallel data only} \\
\hline
en--de & 174,443 & 3,575,407 & 3,353,855 \\
en--he & 187,817 & 3,862,985 & 2,958,136 \\
en--vi & 133,317 & 2,965,962 & 3,361,789 \\
\hline
\multicolumn{4}{c}{IWSLT parallel data + back-translated data} \\
\hline
en--de & 269,213 & 5,843,264 & 5,537,986 \\
en--he & 282,587 & 6,130,842 & 4,728,840 \\
en--vi & 228,087 & 5,413,428 & 6,232,006 \\
\hline
\end{tabular}
\caption{Number of sentences and tokens in the training corpora used in our experiments.}
\label{tab:corpus}
\end{table}

In order to study the domain robustness of our MTL~DA approach, we also evaluated the systems trained on some out-of-domain test sets. We chose
the IT, law and medical test sets released by \citet{muller-etal-2020-domain}\footnote{\url{https://github.com/ZurichNLP/domain-robustness}} and also used by \citet{wang-sennrich-2020-exposure} for English--German.


\textcolor{black}{
Corpora were tokenised and truecased with the Moses scripts;\footnote{\url{https://github.com/moses-smt/mosesdecoder/tree/master/scripts}} then, sentences with more than 100 or less than 5 tokens were removed from the training corpora. Afterwards, byte-pair encoding (BPE) with 10,000 merge operations~\citep{bpe} was applied on the concatenation of the source and target sides of the training corpora to obtain the vocabulary. Finally, those sentence pairs in the training corpora with more than 100 BPE tokens were removed.}

One-to-many word alignments in both translation directions
were obtained using \texttt{mgiza++}~\citep{gao-vogel-2008-parallel,och-ney-2003-systematic}.\footnote{\url{https://github.com/moses-smt/mgiza}} Source-to-target word alignments were used for the \emph{mono} transformations; the one-to-one word alignments required by the \emph{replace} transformation were obtained by computing the intersection between the one-to-many word alignments in both translation directions.
The bilingual lexicon for the \emph{replace} transformation was built by  associating to each source word the target word it is most frequently aligned with in the one-to-one word alignments.


\paragraph{Training.}
Our neural model is a transformer-based model as defined by~\citet{vaswani2017attention}, with the exception of the amount of \emph{warmup\_steps}, which was set to 8,000. All the experiments were carried out on a single GPU with mini-batches made of 4,000 tokens. Validation was done every 1,000 updates, and the patience based on the BLEU score on the development set was set to 6 validation cycles; we then kept the intermediate model performing best on the development set.  
We trained the systems with the \texttt{fairseq} toolkit~\citep{ott-etal-2019-fairseq}. For RAML and SwitchOut, we integrated into \texttt{fairseq} the sampling function released by~\citet{wang-etal-2018-switchout}.\footnote{Code available at \url{https://github.com/transducens/mtl-da-emnlp}.}

Systems trained with MTL~DA were fine-tuned on the main (translation) task after being trained on the combination of the main and auxiliary tasks. 
When combining different auxiliary tasks, a different special token was used for each one. 


\paragraph{DA hyperparameters.}
The proportion of words affected by the \emph{swap}, \emph{token} and \emph{replace} transformations is controlled by a hyperparameter $\alpha$, whereas RAML and SwitchOut are governed by a temperature $\tau$. For each language pair, we explored values of $\alpha$ in $[0.1,0.9]$ at intervals of $0.1$, and values of $\tau$ around the best values reported by~\citet{wang-etal-2018-switchout}.\footnote{$\tau^{-1} \in \{ 0.5, 0.6, 0.7, 0.8, 0.85, 0.9, 0.95, 1.0, 1.1, \\ 1.2, 1.3 \}$}  
The results reported are those obtained with the model that maximizes BLEU on the development set. \textcolor{black}{ The best hyperparameters obtained for the experiments with the IWSLT parallel data were reused for the experiments with the training set extended with back-translated data. }

\section{Results and discussion}
\label{sec:results}

\begin{table*}[tb]
\centering
\fontsize{10.2}{12.0}\selectfont
\begin{tabular}{lcccccccccccc}
\multicolumn{7}{c}{IWSLT parallel data only} \\
\hline
\textbf{Task} & \textbf{en-de} & \textbf{de-en} & \textbf{en-he} & \textbf{he-en} & \textbf{en-vi} & \textbf{vi-en} \\
\hline
\hline
baseline & $24.7 \pm 0.2$ & $30.0 \pm 0.1$ & $21.5 \pm 0.3$ & $32.4 \pm 0.1$ & $28.9 \pm 0.1$ & $27.5 \pm 0.4$ \\
\hline
SwitchOut & $25.3\pm0.2$ & $30.1\pm0.2$ & $21.6\pm0.6$ & $32.1\pm0.4$ & $28.5\pm0.2$ & $27.3\pm0.6$ \\
RAML & $25.4\pm0.2$ & $30.3\pm0.1$ & $21.9\pm0.1$ & $32.1\pm0.1$ & $28.6\pm0.5$ & $27.3\pm0.5$ \\
SwitchOut+RAML & $\mathbf{25.7\pm0.4}$ & $30.3\pm0.5$ & $22.1\pm0.4$ & $32.1\pm0.4$ & $29.1\pm0.4$ & $27.5\pm0.3$ \\
\hline
swap & $25.1 \pm 0.2$ & $30.3 \pm 0.1$ & $22.1 \pm 0.4$ & $32.5 \pm 0.6$ & $28.8 \pm 0.2$ & $28.3 \pm 0.6$ \\ 
token & $25.4 \pm 0.2$ & $30.0 \pm 0.3$ & $21.5 \pm 0.2$ & $32.4 \pm 0.8$ & $29.3 \pm 0.3$ & $28.2 \pm 0.3$ \\
source & $25.3 \pm 0.1$ & $30.1 \pm 0.4$ & $21.5 \pm 0.3$ & $32.7 \pm 0.2$ & $28.9 \pm 0.3$ & $27.6 \pm 0.2$ \\
reverse & $\mathbf{26.1 \pm 0.3}$ & $30.2 \pm 0.1$ & $22.4 \pm 0.2$ & $\mathbf{33.4 \pm 0.3}$ & $29.4 \pm 0.3$ & $28.2 \pm 0.4$ \\
mono & $25.7 \pm 0.1$ & $30.4 \pm 0.2$ & $22.0 \pm 0.1$ & $32.5 \pm 0.6$ & $29.3 \pm 0.4$ & $27.7 \pm 0.4$ \\
replace & $\mathbf{25.8 \pm 0.3}$ & $30.7 \pm 0.2$ & $22.5 \pm 0.2$ & $\mathbf{33.5 \pm 0.3}$ & $29.5 \pm 0.3$ & $28.3 \pm 0.9$ \\
reverse+replace & $\mathbf{26.3 \pm 0.1}$ & $\mathbf{31.1 \pm 0.3}$ & $\mathbf{22.9 \pm 0.2}$ & $\mathbf{33.9 \pm 0.2}$ & $\mathbf{30.1 \pm 0.5}$ & $28.8 \pm 0.2$ \\ 
reverse+mono+replace & $\mathbf{26.4 \pm 0.6}$ & $\mathbf{31.4 \pm 0.3}$ & $\mathbf{23.2 \pm 0.3}$ & $\mathbf{33.9 \pm 0.5}$ & $\mathbf{30.5 \pm 0.2}$ & $\mathbf{29.4 \pm 0.3}$ \\
\hline 
\end{tabular}
\caption{Mean and standard deviation of the BLEU scores obtained when translating in-domain test sets with the baseline system, three other reference systems, and our MTL~DA approach, using different auxiliary tasks and combinations of them. Systems were trained only on IWSLT parallel data.
The best results for each language pair, and those falling within one standard deviation from them, are highlighted in bold.}
\label{tab:results-train-bleu}
\end{table*}

\begin{table*}[tb]
\centering
\fontsize{10.2}{12.0}\selectfont
\begin{tabular}{lccccccccc}
\multicolumn{4}{c}{IWSLT parallel data + back-translated data} \\
\hline
\textbf{Task} & \textbf{de-en}  & \textbf{he-en} &  \textbf{vi-en} \\
\hline
\hline
baseline & $31.3 \pm 0.5$ & $34.5 \pm 0.1$ & $29.3 \pm 0.3$ \\
SwitchOut+RAML & $\mathbf{31.7\pm0.8}$ & $34.1\pm0.7$ & $\mathbf{29.9\pm0.5}$ \\
reverse+mono+replace &  $\mathbf{32.3 \pm 0.1}$ & $\mathbf{35.3 \pm 0.3}$ & $\mathbf{30.4 \pm 0.7}$ \\
\hline
\end{tabular}
\caption{Mean and standard deviation of the BLEU scores obtained when translating in-domain test sets with the baseline system, the combination of SwitchOut and RAML, and the best combination of auxiliary task in our MTL~DA approach. Systems were trained on a combination of parallel and back-translated data.
The best results for each language pair, and those falling within one standard deviation from them, are highlighted in bold.}
\label{tab:results-train-bleu-backt}
\end{table*}

\paragraph{IWSLT parallel data.}
\textcolor{black}{Table~\ref{tab:results-train-bleu} reports the mean and standard deviation of the translation performance, measured in terms of BLEU~\citep{papineni},\footnote{\texttt{sacrebleu}~\citep{post-2018-call} version string: \texttt{BLEU+case.mixed+lang.vi-en+numrefs.1+} \texttt{smooth.exp+tok.13a+version.1.5.0}} of three different executions for each of the systems trained on the IWSLT parallel data. chrF++ scores~\citep{popovicchrfpp}, that show the same trend, are available in Appendix~\ref{app:chrf}.} 
The results show that our MTL~DA approach consistently outperforms the baseline system in all language pairs and translation directions. In general, the auxiliary tasks
\emph{reverse} (translation into the target language but in the reverse order) and \emph{replace} (random replacement of target words and the source words they are aligned with) are the best performing ones. 
\emph{swap} (random swapping of words) and \emph{source} (copying the source sentence) often perform worse than the former tasks, which suggests that a non-systematic word order or a completely different vocabulary in the target could negatively influence the main task.

Interestingly, using the three best auxiliary tasks together further improves the performance, achieving the best results in all translation tasks with BLEU scores between 1.1 and 1.9 points over the baseline.\footnote{\textcolor{black}{Even though MTL DA is intended for low-resource language pairs, we conducted preliminary experiments on large training data using the English--German WMT 2014 dataset~\citep{gao2019soft}, which contains around~4.5M training parallel sentences. The results show a gain of around 1~BLEU point over the baseline for German--English and the same performance for English--German. In any case, the performance of MTL DA on large data sets remains to be studied.
}} This suggests that different auxiliary tasks affect the encoder in different ways and are somehow complementary.

A comparison of our MTL~DA approach with RAML, SwitchOut and their combination (SwitchOut+RAML) shows that our approach, being much simpler in nature, also outperforms them.

\paragraph{Back-translated data. }
\textcolor{black}{
Table~\ref{tab:results-train-bleu-backt} shows the results obtained with the training set extended with back-translated data. In these experiments, we only evaluated MTL~DA with the best performing combination of auxiliary tasks. As for SwitchOut and RAML, we only evaluated their combination, which according to Table~\ref{tab:results-train-bleu} performs better than any method in isolation. Although the differences are slightly smaller, we can observe the same trend in the results: MTL~DA still outperforms the baseline and the combination of SwitchOut and RAML. In addition, the results show that MTL~DA and back-translation are two complementary DA approaches.
}


\paragraph{Domain robustness.}

\begin{table*}[tb]
\centering
\fontsize{10.3}{12.0}\selectfont
\begin{tabular}{lcccccccccccc}
\textbf{Domain} & \multicolumn{2}{c}{IT} & \multicolumn{2}{c}{Law} & \multicolumn{2}{c}{Medical} \\
\hline
\textbf{Direction} & \textbf{en-de} & \textbf{de-en} & \textbf{en-de} & \textbf{de-en} & \textbf{en-de} & \textbf{de-en} \\
\hline 
\hline
baseline & $3.0\pm0.3$ & $6.2\pm1.9$  & $6.0\pm0.7$ & $8.1\pm0.8$ & $9.5\pm0.6$  & $10.7\pm1.5$ \\
\hline
SwitchOut & $5.1\pm0.8$ & $5.3\pm2.5$  & $7.6\pm0.2$ & $7.8\pm0.5$ & $11.8\pm0.2$ & $10.1\pm1.2$ \\
RAML & $5.0\pm1.4$ & $6.3\pm3.1$  & $7.8\pm0.3$ &$\mathbf{8.5\pm0.7}$ & $11.7\pm1.0$ & $11.6\pm1.6$  \\
SwitchOut+RAML & $8.0\pm0.5$ & $6.8\pm0.7$& $7.8\pm0.2$ & $7.5\pm1.2$ & $12.5\pm1.1$ & $10.1\pm1.0$  \\
\hline
reverse & $10.2\pm0.6$ & $\mathbf{10.7\pm0.6}$ &  $7.6\pm0.4$ & $8.0\pm0.6$ & $12.8\pm0.4$ &  $12.1\pm0.5$ \\
reverse+mono+replace & $\mathbf{13.2\pm1.5}$ & $\mathbf{11.3\pm0.9}$  & $\mathbf{10.2\pm0.4}$ & $\mathbf{9.9\pm0.8}$ & $\mathbf{15.9\pm1.0}$ & $\mathbf{14.3\pm0.8}$  \\
\hline
\end{tabular}
\caption{Mean and standard deviation of the BLEU scores obtained when translating out-of-domain texts in the IT, law and medical domains. The best results for each language pair, and those falling within one standard deviation from them, are highlighted in bold.}
\label{tab:results-outdomain-bleu}
\end{table*}

Concerning the out-of-domain evaluation, Table \ref{tab:results-outdomain-bleu} shows the BLEU scores obtained by each 
system; chrF++ scores show the same trend and are provided in Appendix~\ref{app:chrf}. We restricted the MTL~DA evaluation to the \emph{reverse} task and the combination of the best three auxiliary tasks, \textcolor{black}{and only report the results obtained with systems trained on the IWSLT parallel data.}

As can be seen, MTL~DA outperforms the baseline system and 
RAML/SwitchOut. Even the \emph{reverse} auxiliary task, which does not modify the vocabulary of the target sentence in any way and does not add infrequent words to the training corpus, enhances the domain robustness of the system. 


\section{Explainability}
\label{sec:explain}


To confirm that the systematic improvements in translation quality and enhanced domain robustness are related to the encoder being exposed during training to more situations where a good source representation is crucial, we carried out an analysis of the relative source and target contributions to the generation decisions of the NMT system. According to \citet{voita2020analyzing}, systems trained with more data tend to rely more on source information; we expect MTL~DA to produce the same effect. 


Another aspect that will account for the positive impact of MTL~DA in the system's encoder is the generation of \emph{hallucinations}~\citep{hallucinations2018}: 
completely inadequate translations that usually occur under domain shift~\citep{muller-etal-2020-domain}, due to the system relying too much on the target context~\citep{voita2020analyzing}. We expect systems trained with MTL~DA to produce less hallucinations. To validate this last hypothesis, we carried out an hallucination analysis on the results of the domain shift experiments.

\paragraph{Relative source and target contributions.}

\begin{figure*}
     \centering
     \begin{subfigure}[b]{0.30\textwidth} 
         \includegraphics[width=\textwidth]{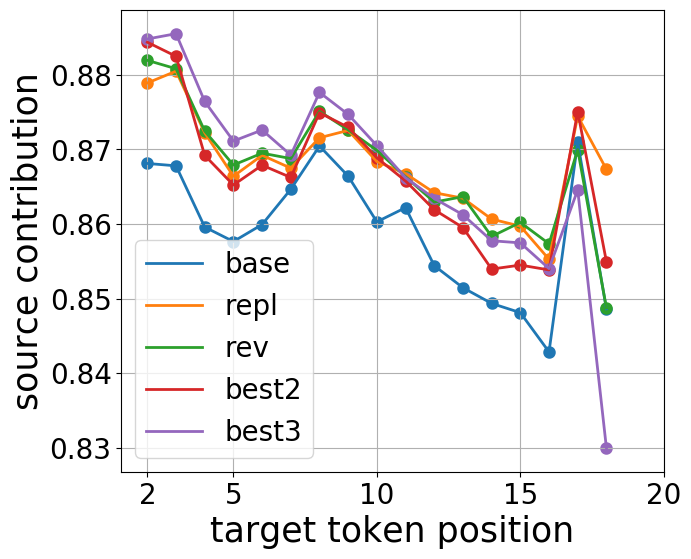}
         \caption{English--German}
     \end{subfigure}
    \hfill
    \begin{subfigure}[b]{0.30\textwidth}
         \includegraphics[width=\textwidth]{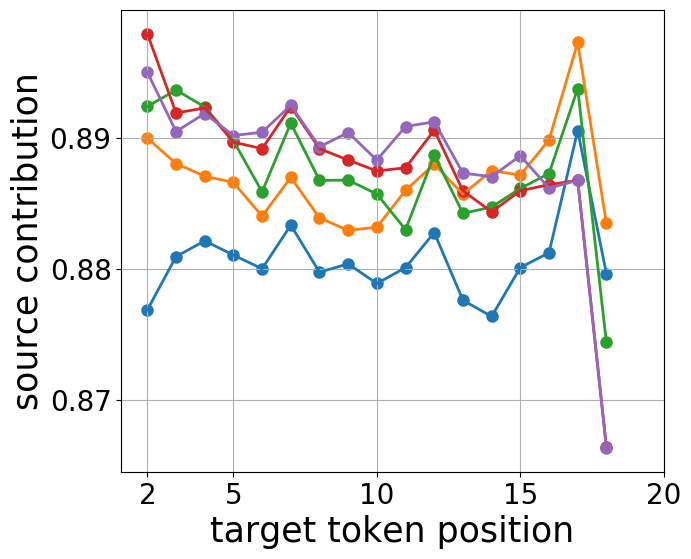}
         \caption{English--Hebrew}
     \end{subfigure}
    \hfill 
         \begin{subfigure}[b]{0.30\textwidth}
         \includegraphics[width=\textwidth]{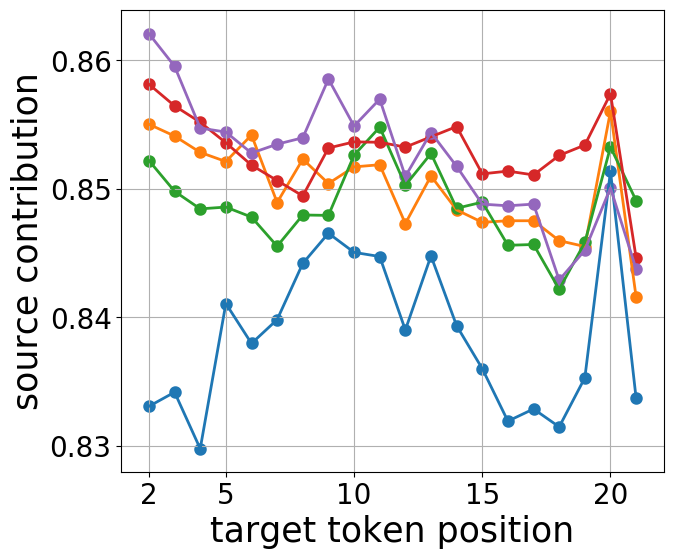}
         \caption{English--Vietnamese}
     \end{subfigure}
     \hfill
     \begin{subfigure}[b]{0.30\textwidth}
         \includegraphics[width=\textwidth]{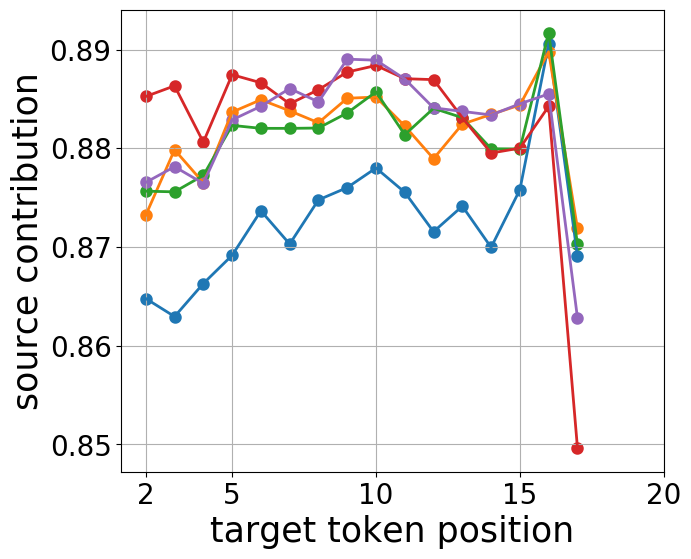}
         \caption{German--English}
     \end{subfigure}
    \hfill
     \begin{subfigure}[b]{0.30\textwidth}
         \includegraphics[width=\textwidth]{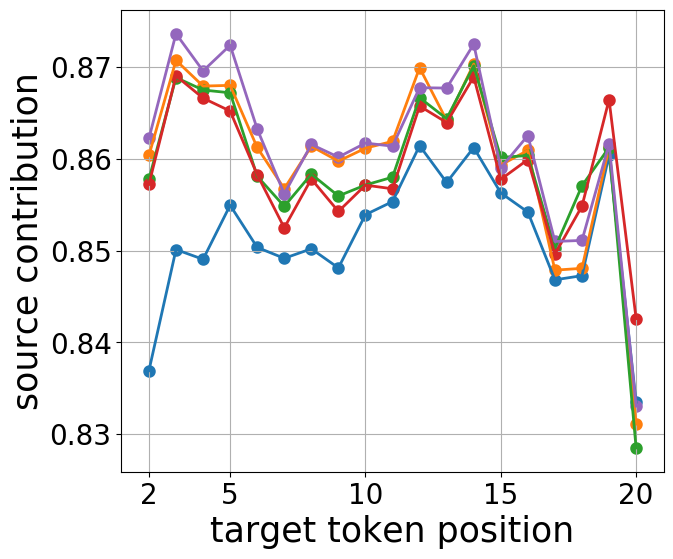}
         \caption{Hebrew--English}
     \end{subfigure}
   \hfill 
     \begin{subfigure}[b]{0.30\textwidth}
         \includegraphics[width=\textwidth]{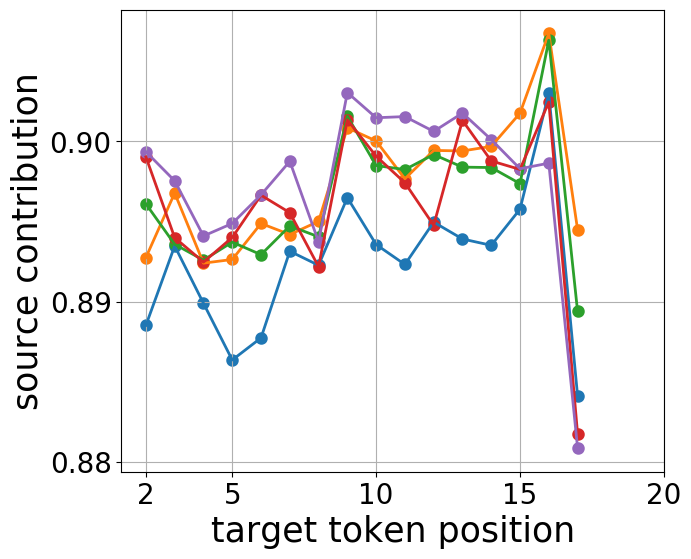}
         \caption{Vietnamese--English}
     \end{subfigure}
\caption{For each translation task, total contribution of the source tokens $R_t(\mathbf{x})$ to the production of the target tokens. \emph{best2} stands for the combination of the two best auxiliary tasks within our MTL~DA framework: \emph{reverse} and \emph{replace}. \emph{best3} stands for the combination of the three best ones:     \emph{reverse}, \emph{replace} and \emph{mono}.}
\label{fig:source_influence}     
\end{figure*}

We used \emph{layer-wise relevance propagation} (LRP), as adapted to transformers by~\citet{voita2020analyzing}, to compute the relative contribution of source and target tokens to each prediction made by the system. LRP allows us to compute $R_t(x_i)$ and $R_t(y_j)$, the relative contribution of source token $x_i$ and target token $y_j$, respectively, to the prediction emitted by the network at time $t$. The total relevance at each time step is $1$, i.e. for all time steps $t$ the following equation holds, where $R_t(\mathbf{x})$ and $R_t(\mathbf{y})$ stand for the total contribution of the source tokens and target tokens, respectively:

$$
\sum_i R_t(x_i) + \sum_j R_t(y_j) = R_t(\mathbf{x}) + R_t(\mathbf{y}) = 1
$$

To have reliable comparisons, we also follow \citet{voita2020analyzing} and evaluate the relative source and target contributions on a subset of sentences from a held-out corpus with the same source length and the same target length. In this way, we can fairly compare different strategies, since we teacher-force the reference translations when computing LRP so as to obtain translations with exactly the same length. The held-out corpus used is the concatenation of all the development corpora released for the corresponding IWSLT task,\footnote{English--German: dev2010, dev2012, tst2010, tst2011, tst2012, tst2013, tst2014; English--Hebrew: dev2010, tst2010, tst2011, tst2012, tst2013, tst2014; English--Vietnamese: dev2010, tst2010, tst2011, tst2012, tst2013. } while the subset chosen is the largest set of parallel sentences with the same source length and the same target length, as long as there are at least 16 tokens in each side.\footnote{It contains 48 en--de sentences (768 en tokens and 816 de tokens), 28 en--he sentences (448 en tokens and 560 he tokens), and 28 en--vi sentences (448 en tokens and 560 vi tokens). Similar trends in source contribution were observed when selecting a larger subset with shorter sentences. } 
To compute the relative contributions, we retrained 
the baseline and the MTL~DA systems featuring the best performing auxiliary tasks \textcolor{black}{on the IWSLT parallel data} with the toolkit released by \citet{voita2020analyzing}.\footnote{\url{https://github.com/lena-voita/the-story-of-heads}}
Figure~\ref{fig:source_influence} depicts the total source contribution at each generation time step $t$ for the different translation tasks and systems. We skip the first time step, when there is no target prefix available, and show the source contribution for the EOS token too. For MTL~DA models, the source contains the special token corresponding to the main task.

For the translation tasks with English as source language, we can observe the same trends as~\citet{voita2020analyzing}: source influence decreases as decoding progresses. There is also a peak in the penultimate token, which may express that the decoder is checking whether it has translated all the content from the source sentence before emitting the full stop at the end of the sentence. 
When English is the target language, plots are flatter: source influence does not decrease as decoder advances.  
The fact that English grammar is simpler, lacking gender and case agreements, could explain that the decoder needs to check previous tokens less.
 
These results confirm the utility of MTL~DA: the baseline system is systematically the one where the source has the smallest influence, and auxiliary tasks increase source influence in all translation tasks. 
Differences are larger at the beginning of decoding, but remain throughout the sentence.
MTL~DA achieves, only with artificially augmented data, an increase in source influence comparable to that reported by~\citet[Fig. 6]{voita2020analyzing} when the size of genuine parallel data increases.

Finally, no consistent differences in source influence could be found between the \emph{reverse} and \emph{replace} auxiliary tasks. The systems combining multiple auxiliary tasks, however, are consistently the ones with the highest source influence, thus confirming the complementarity of the tasks.


\paragraph{Hallucinations.}
To estimate the number of hallucinations produced by the systems evaluated, we follow the procedure proposed by~\citet{hallucinations2018} and used by~\citet{curious2021}. Although their interest was in detecting those sentences that induced the generation of hallucinations 
after introducing spurious tokens in the input, we adapted it to automatically measure the number of input sentences in a test set for which the corresponding output seems to be an hallucination. To this end, we use an adjusted version of BLEU which only takes into account the precision of unigrams and bigrams with weights $0.8$ and $0.2$, respectively, as proposed by~\citet{hallucinations2018}. If the sentence-level adjusted BLEU of the lowercased emitted translation is below a certain threshold ($10$ in our experiments), 
it is taken as a sign of hallucination.

\begin{figure}[tb]
     \centering
     \begin{subfigure}[b]{0.505\textwidth} 
\includegraphics[width=0.96\textwidth]{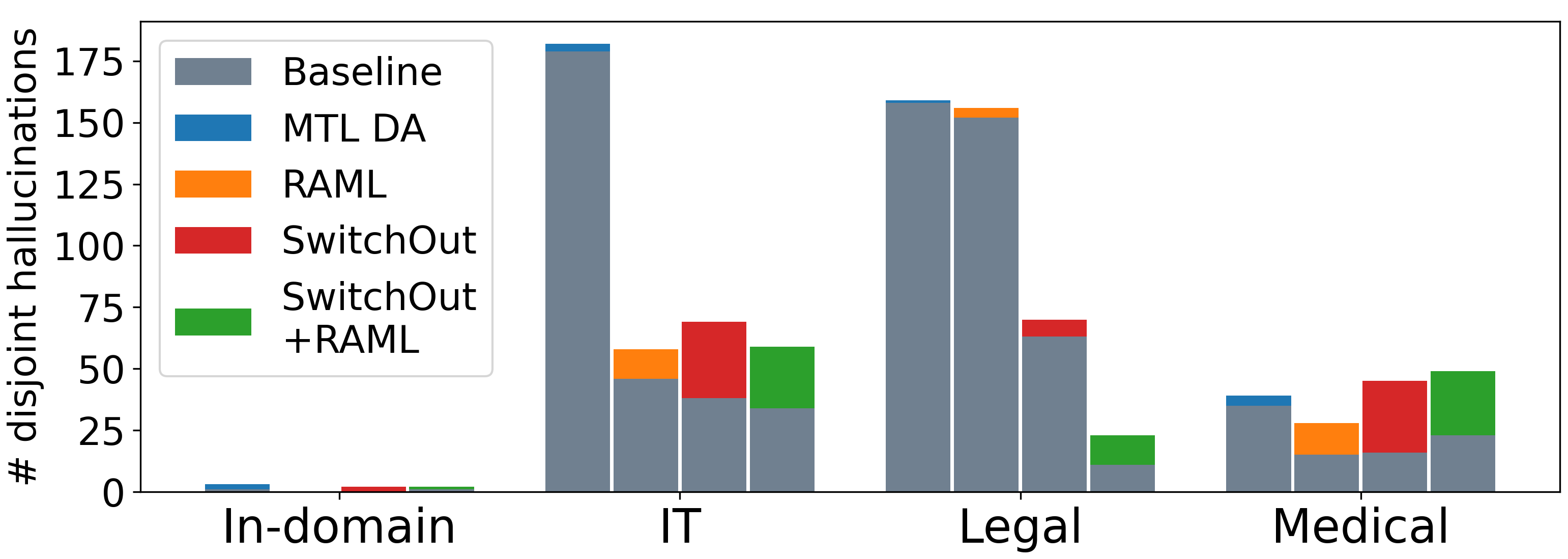}
         \caption{English--German}
     \end{subfigure}
     \begin{subfigure}[b]{0.505\textwidth}
\includegraphics[width=0.96\textwidth]{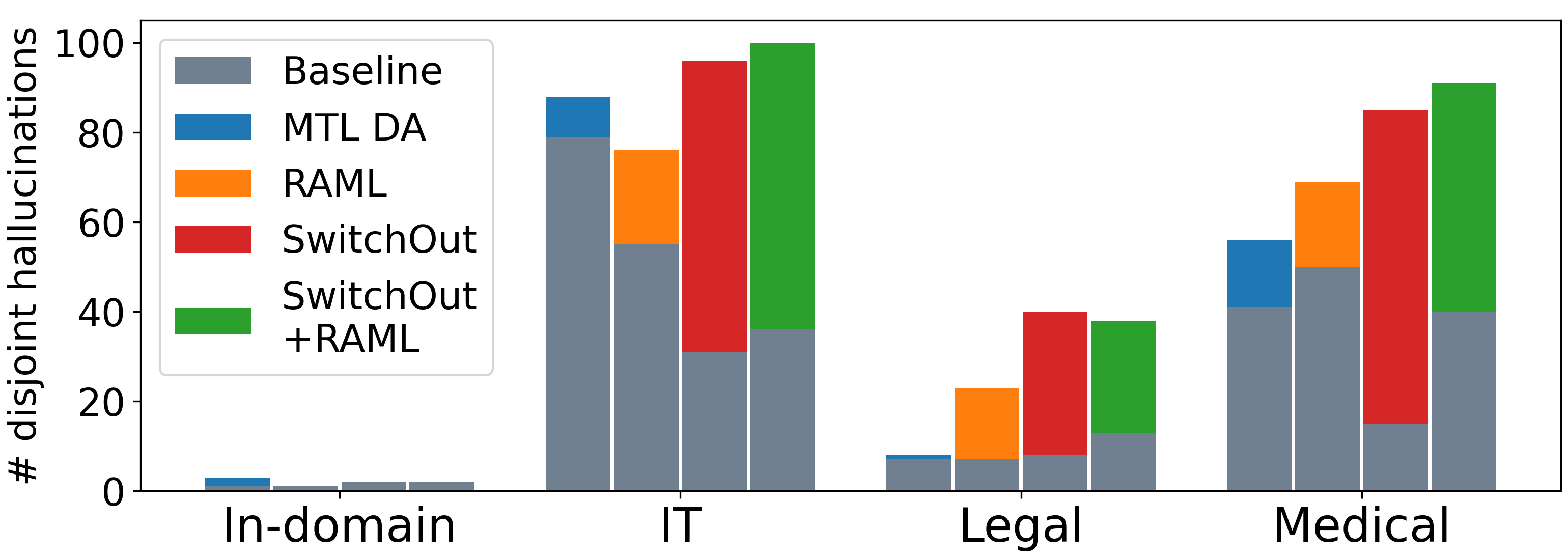}
         \caption{German--English}
     \end{subfigure}
\caption{Number of disjoint hallucinations of the baseline system (in grey) and the systems trained with different DA methods. Our MTL~DA approach (in blue) corresponds to the system labeled as \emph{reverse+mono+replace} in Table~\ref{tab:results-train-bleu}.}
\label{fig:hallucinations}     
\end{figure}

We evaluate the tendency to produce hallucination of the baseline as compared to our MTL~DA approach combining the auxiliary tasks \emph{reverse}, \emph{mono} and \emph{replace}, and to the SwitchOut, RAML and SwitchOut+RAML systems. Sentences whose translations cannot be regarded as hallucinations are not relevant to our study, neither are those for which the baseline and the system to which it is compared to both hallucinate. 
We therefore count the number of sentences that induce an hallucination on one of the systems but not on the other.\footnote{As an example, the evaluation corpus contains the pair (``To Select an Object'', ``Objekt auswählen''). The German--English baseline produces ``Ein Objekt auszuwählen, um ein Objekt auszuwählen.'' which shows an hallucination in the form of a repetition, whereas the MTL~DA method gives a better ``Um ein Objekt auszuwählen.''.} We consider that the other system is not hallucinating if its adjusted BLEU is at least 20 BLEU points higher. Fig.~\ref{fig:hallucinations} represents these data for the English--German translation tasks for the same domains (in-domain, IT, legal and medical) and corpora used for the 
domain robustness evaluation reported in Sec.~\ref{sec:results}. The grey bar represents the number of hallucinations of the baseline that are not labeled as such by the corresponding DA system; consequently, higher values are better. Additionally, the color bars represent the number of hallucinations of the DA system that cannot be labeled as such in the baseline system (shorter bars are better).

As can be seen, disjoint hallucinations barely happen with in-domain data, but they can be easily found when considering out-of-domain data. In every domain and in both translation directions, the blue bar (the one representing our MTL~DA method) is clearly the shortest one in almost all cases, while at the same time the grey bar is the largest one. This is a clear sign of reduction of hallucinations in the systems trained with MTL~DA. Appendix~\ref{app:hallucina} presents supplementary details on the BLEU thresholds and the smoothing technique considered when computing the scores.

\section{Related work}
\label{sec:related}

The back-translation~\citep{sennrich2016edinburgh} approach for leveraging additional target monolingual data to produce additional training samples is probably the most popular DA approach for NMT. The set of related approaches covered in this section, however, mainly focus on methods that, as MTL~DA, do not require additional resources besides the training parallel corpus. 

\citet{li-etal-2019-understanding-data} evaluate back- and forward-translation in such a setting. They train forward and backward NMT systems on the available parallel data and use them to produce new synthetic samples by translating either the target side~\citep{sennrich2016edinburgh} or the source side~\citep{zhang-zong-2016-exploiting} of the original training corpus.


The approaches we have evaluated in our experiments, RAML~\citep{norouzi2016reward} and its extension to the source language, SwitchOut~\citep{wang-etal-2018-switchout}, aim at extending the support of the empirical data distribution and keeping it smooth (similar sentence pairs have similar probabilities). To that end, they replace words with other words sampled from a uniform distribution over the vocabulary, which, in practice, results in infrequent words being overrepresented. \citet{guo-etal-2020-sequence} presented a related approach to encourage compositional behaviour: replaced words are selected from another sentence and not from the vocabulary.   


Some of our auxiliary tasks have already been used for DA, but mostly on the source side and rarely in an MTL framework.  Replacing tokens with placeholders (as we do in \emph{token}) has already been applied by~\citet{zhang2020token} to the source language, in combination with 
auxiliary tasks involving detecting replaced and dropped tokens. \citet{xie2017data} also evaluate the impact of replacements on the target 
data, but do not follow an MTL approach.
Word dropout
~\citep{sennrich2016edinburgh,NIPS2016_076a0c97,shen2020simple} can also be considered a related approach.

Regarding changes to word order, in addition to the proposals by
\citet{artetxe2017unsupervised} and \citet{lample2018unsupervised}, it is worth highlighting the strategy proposed by \citet{zhang2019regularizing} who apply a self-translation approach using a right-to-left decoder. Unlike our MTL~DA framework, they need to generate translations from the model during training and adjust multiple terms in the training loss.


There are more DA approaches based on replacing words which are worth mentioning.
\citet{xie2017data} randomly replace words in the source side of the training samples. 
\citet{gao2019soft} replace words selected at random with \emph{soft words} whose representations are obtained from the probability distribution provided by a language model. 
\citet{Fadaee_2017} replace a number of words in their training samples by infrequent words in order to improve the performance of the NMT model when dealing with them at translation time. 
Words to be replaced are identified using
a large source language model. 
Once the source words to be replaced are identified, a word-alignment model and a probabilistic dictionary are used to also replace the corresponding counterpart by the most probable translation of the new source word. In our MTL~DA framework, the \emph{replace} transformation, which is similar to \citet{Fadaee_2017}'s work, does not require any language model.  
 
Regarding back-translation, \citet{edunov-etal-2018-understanding} apply several simple transformations (word deletion, replacement, swapping) to back-translated data reporting a noticeable improvement. 
In relation with the special token we use to prevent negative transfer between tasks, \citet{caswell2019tagged} propose a similar strategy to identify synthetic samples when combining actual parallel data and back-translated data for training. \citet{YANG2019240} extends this work by including forward-translated data for training using two different special tokens to distinguish the two types of synthetic data. 

\section{Concluding remarks}\label{sec:conclusions}

In this paper, we have presented a multi-task learning approach for data augmentation (MTL~DA) in NMT. We deviate from common approaches that aim at extending the support of the empirical data distribution by generating new samples that 
are likely under such distribution. We propose instead to carry out DA in a MTL manner, by artificially generating new sentence pairs with aggressive transformations, such as reversing the order of the target sentence, 
which may make the target sentence completely unfluent. Translating into these augmented sentences constitute new tasks that provide new contexts during training where the target prefix is not informative enough to predict the next word, thus strengthening the encoder and forcing the system to rely more on it.

Experiments carried out on six low-resource translation tasks that usually serve as benchmark for DA show consistent improvements over a baseline system (on average around 1.6 BLEU points) and over strong DA methods that aim at extending the support of the empirical data distribution without MTL. Moreover, additional analyses show that the systems trained with MTL~DA rely more on the source tokens, are more robust against domain shift and suffer less hallucinations.

MTL~DA is agnostic to the NMT model architecture and does not require elaborated preprocessing steps, training additional systems, or data besides the available training parallel corpora. 
Furthermore, it could be combined with existing DA methods, \textcolor{black}{in addition to back-translation}, specially those that operate on the source side~\citep{wang-etal-2018-switchout,gao2019soft}, since our transformations mainly address the target. 

We expect this strategy to inspire the implementation of new auxiliary tasks to be used for MTL~DA, specially those aimed at 
improving the training dynamics of the system. We believe that further improvements could be obtained by following more elaborated strategies for multi-task learning, such as changing the proportion of data for the different tasks, evaluating different ways of parameter sharing between the different tasks (e.g. sharing the encoder but not the decoder), and using other training schedules~\citep{pmlr-v80-chen18a}. Finally, we conclude that making the encoder representation essential to minimize the loss during training should be embraced as a potential way of boosting NMT quality.

\section*{Acknowledgments}
We thank Wilker Aziz, a collaboration with whom served as an inspiration for this work, and Ivan Titov, who gave us valuable hints about the role of encoder representations. Work funded by the European Union’s Horizon 2020 research and innovation programme under grant agreement number 825299, project Global Under-Resourced Media Translation (GoURMET); and by Generalitat Valenciana through project GV/2021/064. The computational resources used for the experiments were funded by the European Regional Development Fund through project IDIFEDER/2020/003.

\bibliography{anthology,emnlp2020}
\bibliographystyle{acl_natbib}

\clearpage
\appendix
\label{sec:appendix}

\section{Training details}\label{sec:train-details}


\begin{table*}[bth]
\centering
\fontsize{10.2}{12.0}\selectfont
\begin{tabular}{llcccccccccccc}
\textbf{Hyperparam.} & \textbf{Task} & \textbf{en-de} & \textbf{de-en} & \textbf{en-he} & \textbf{he-en} & \textbf{en-vi} & \textbf{vi-en} \\
\hline \hline
$\tau_x^{-1}$ & SwitchOut & $1.1$ & $0.85$ & $0.9$ & $0.6$ & $0.95$  & $1.0$ \\
$\tau_y^{-1}$  & RAML & $0.7$ & $1.1$ & $0.5$ & $0.95$ & $0.85$ & $0.8$ \\
$\tau_y^{-1}$  & SwitchOut+RAML & $0.7$ & $1.0$  & $0.95$  & $1.0$  & $0.5$ & $0.8$ \\
\hline
$\alpha$  & swap & $0.2$ & $0.1$  & $0.4$  & $0.2$  & $0.2$ & $0.1$ \\
$\alpha$  & token & $0.8$ & $0.8$ & $0.1$ & $0.7$ & $0.3$ & $0.6$ \\
$\alpha$ & replace & $0.2$ & $0.5$ & $0.3$ & $0.2$ & $0.3$ & $0.2$ \\ 
\hline
\end{tabular}
\caption{Data augmentation hyperparameters that maximized BLEU on the development set.}
\label{tab:hyperparams}
\end{table*}

The DA hyperparameters that maximized BLEU on the development set on the first training run are depicted in Table~\ref{tab:hyperparams}. Subsequent training runs were executed only with these best values. For the combination of SwitchOut and RAML, following~\citet{wang-etal-2018-switchout}, firstly the best $\tau_x$ for SwitchOut was determined and, afterwards, the best $\tau_y$ for RAML was sought by fixing $\tau_x$. For Switchout/RAML, the table depicts the value of $\tau^{-1}$, while for the MTL~DA tasks, it shows the value of $\alpha$.

\section{Results with chrF++}
\label{app:chrf}
\textcolor{black}{
Tables~\ref{tab:results-train-chrf} and \ref{tab:results-train-chrf-backt} show the chrF++ scores~\citep{popovicchrfpp} for the in-domain automatic evaluations based on IWSLT parallel training data and on extended training data with back-translation, respectively; they are the counterpart to tables~\ref{tab:results-train-bleu} and~\ref{tab:results-train-bleu-backt}.
Table~\ref{tab:results-outdomain-chrf} shows the chrF++ scores for the
out-of-domain automatic evaluations whose corresponding BLEU scores are reported in Table~\ref{tab:results-outdomain-bleu}. Scores were computed with \texttt{SacreBLEU}~\citep{post-2018-call}.\footnote{Version string: \texttt{chrF2+lang.vi-en+numchars.6+} \texttt{space.false+version.1.5.0}}
}

\begin{table*}[tb]
\centering
\fontsize{10.2}{12.0}\selectfont
\begin{tabular}{lcccccccccccc}
\multicolumn{7}{c}{IWSLT parallel data only} \\
\hline
\textbf{Task} & \textbf{en-de} & \textbf{de-en} & \textbf{en-he} & \textbf{he-en} & \textbf{en-vi} & \textbf{vi-en} \\
\hline \hline
baseline & $52.0 \pm 0.1$ & $53.1 \pm 0.2$ & $47.7 \pm 0.2$ & $54.6 \pm 0.1$ & $48.6 \pm 0.3$ & $49.2 \pm 0.1$ & \\
\hline
SwitchOut &$52.2\pm0.4$ & $53.1\pm0.2$ & $47.6\pm0.4$ & $54.5\pm0.3$ & $48.1\pm0.2$  & $49.7\pm0.6$ \\
RAML & $52.5\pm0.2$ & $53.3\pm0.2$ & $48.2\pm0.3$ & $54.6\pm0.1$ & $48.7\pm0.1$ & $49.8\pm0.3$ \\
SwitchOut+RAML & $52.5\pm0.1$ &  $53.3\pm0.5$ &  $48.0\pm0.1$ & $54.6\pm0.3$ & $48.7\pm0.3$ & $49.6\pm0.5$ \\
\hline
swap & $52.4 \pm 0.1$ & $53.5 \pm 0.1$ & $48.1 \pm 0.2$ & $55.1 \pm 0.2$ & $48.5 \pm 0.1$ & $50.3 \pm 0.3$ \\
token & $52.6 \pm 0.4$ & $53.1 \pm 0.4$ & $47.7 \pm 0.3$ & $54.7 \pm 0.4$ & $48.9 \pm 0.2$ & $50.2 \pm 0.1$ \\
source & $52.4 \pm 0.2$ & $53.6 \pm 0.4$ & $47.5 \pm 0.3$ & $54.9 \pm 0.3$ & $49.2 \pm 0.2$ & $49.9 \pm 0.4$ \\
reverse & $53.1 \pm 0.3$ & $53.7 \pm 0.2$ & $48.4 \pm 0.2$ & $55.5 \pm 0.1$ & $49.1 \pm 0.5$ & $50.6 \pm 0.0$ \\
mono & $52.6 \pm 0.3$ & $53.8 \pm 0.3$ & $48.1 \pm 0.1$ & $54.9 \pm 0.4$ & $49.0 \pm 0.1$ & $49.9 \pm 0.2$ \\
replace & $\mathbf{53.4 \pm 0.2}$ & $54.2 \pm 0.2$ & $48.6 \pm 0.1$ & $\mathbf{55.8 \pm 0.4}$ & $49.5 \pm 0.2$ & $\mathbf{50.7 \pm 0.5}$ \\ 
reverse+replace & $\mathbf{53.7 \pm 0.4}$ & $\mathbf{54.5 \pm 0.3}$ & $\mathbf{49.3 \pm 0.2}$ & $\mathbf{56.1 \pm 0.3}$ & $\mathbf{49.9 \pm 0.4}$ & $\mathbf{51.2 \pm 0.2}$ \\
reverse+mono+replace & $\mathbf{53.8 \pm 0.3}$ & $\mathbf{54.8 \pm 0.2}$ & $\mathbf{49.3 \pm 0.2}$ & $\mathbf{56.1 \pm 0.3}$ & $\mathbf{50.0 \pm 0.2}$ & $\mathbf{51.5 \pm 0.4}$ \\
\hline
\end{tabular}
\caption{
Mean and standard deviation of the chrF++ scores obtained when translating in-domain test sets with the baseline system, three other reference systems, and our MTL~DA approach, using different auxiliary tasks and combinations of them. Systems were trained only on IWSLT parallel data.
The best results for each language pair, and those falling within one standard deviation from them, are highlighted in bold.
}
\label{tab:results-train-chrf}
\end{table*}

\begin{table*}[tb]
\centering
\fontsize{10.2}{12.0}\selectfont
\begin{tabular}{lccccccccc}
\multicolumn{4}{c}{IWSLT parallel data + back-translated data} \\
\hline
\textbf{Task} & \textbf{de-en}  & \textbf{he-en} &  \textbf{vi-en} \\
\hline
\hline
baseline & $54.7 \pm 0.4$ & $56.6 \pm 0.0$ & $51.7 \pm 0.2$ \\
SwitchOut+RAML & $54.9\pm0.6$ & $56.3\pm0.2$ & $\mathbf{52.2\pm0.1}$ \\
reverse+mono+replace &  $\mathbf{55.8 \pm 0.1}$ & $\mathbf{57.4 \pm 0.2}$ & $\mathbf{52.6 \pm 0.7}$ \\
\hline
\end{tabular}
\caption{Mean and standard deviation of the chrF++ scores obtained when translating in-domain test sets with the baseline system, the combination of SwitchOut and RAML, and the best combination of auxiliary task in our MTL~DA approach. Systems were trained on a combination of parallel and back-translated data.
The best results for each language pair, and those falling within one standard deviation from them, are highlighted in bold.}
\label{tab:results-train-chrf-backt}
\end{table*}

\begin{table*}[tb]
\centering
\fontsize{10.3}{12.0}\selectfont
\begin{tabular}{lcccccccccccc}
\textbf{Domain} & \multicolumn{2}{c}{IT} & \multicolumn{2}{c}{Law} & \multicolumn{2}{c}{Medical} \\
\hline
\textbf{Direction} & \textbf{en-de} & \textbf{de-en} & \textbf{en-de} & \textbf{de-en} & \textbf{en-de} & \textbf{de-en} \\
\hline 
\hline
baseline & $25.4\pm2.7$ & $31.6\pm4.4$  & $33.4\pm0.8$ & $33.7\pm0.4$ & $34.6\pm0.5$ & $34.5\pm0.8$ \\
\hline
SwitchOut & $29.7\pm3.2$ & $29.7\pm4.8$  & $34.0\pm0.4$ & $32.8\pm0.8$ & $35.6\pm0.7$ & $34.3\pm0.4$ \\
RAML & $31.3\pm1.8$ & $31.8\pm4.2$ & $34.6\pm0.5$ & $34.2\pm0.1$ & $35.6\pm0.6$ &  $36.0\pm0.6$ \\
SwitchOut+RAML & $33.3\pm0.6$ & $32.8\pm3.1$ & $34.1\pm0.3$ & $33.1\pm0.6$ &$36.3\pm0.9$ & $34.7\pm0.4$\\
\hline
reverse & $35.1\pm0.4$  & $38.2\pm0.2$  & $34.6\pm0.4$  &  $33.7\pm0.6$ & $36.9\pm0.1$ & $36.1\pm0.3$ \\
reverse+mono+replace & $\mathbf{37.6\pm0.3}$ & $\mathbf{39.7\pm0.7}$  & $\mathbf{36.6\pm0.5}$ &$\mathbf{35.5\pm0.8}$  & $\mathbf{39.4\pm0.8}$  & $\mathbf{38.5\pm0.4}$  \\
\hline
\end{tabular}
\caption{Mean and standard deviation of the chrF++ scores obtained when translating out-of-domain texts in the IT, law and medical domains. The best results for each language pair, and those falling within one standard deviation from them, are highlighted in bold.}
\label{tab:results-outdomain-chrf}
\end{table*}

\section{Hallucinations}
\label{app:hallucina}

\begin{figure*}[tb]
     \centering
     \begin{subfigure}[b]{0.48\textwidth} 
         \includegraphics[width=\textwidth]{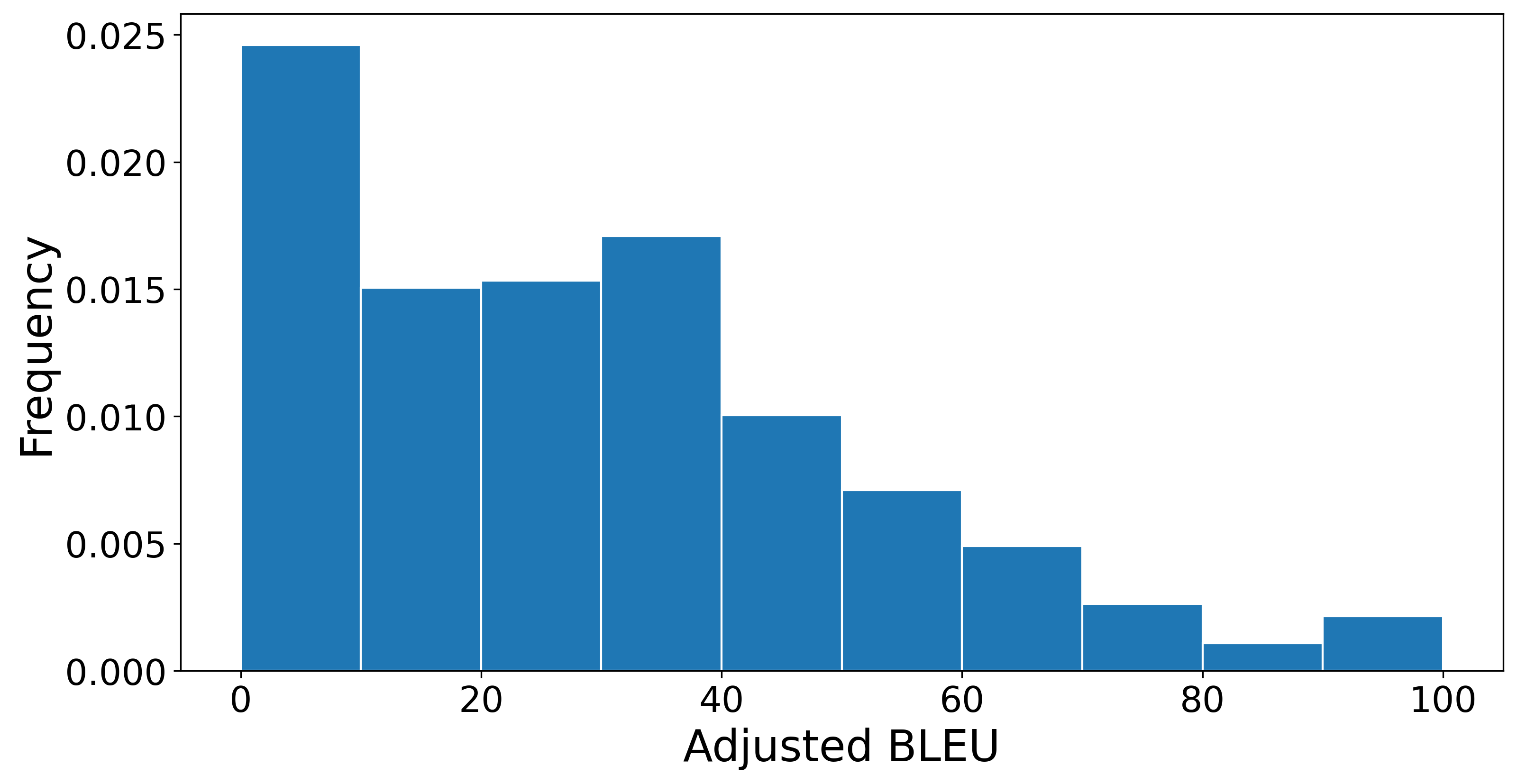}
         \caption{English--German}
     \end{subfigure} 
     \begin{subfigure}[b]{0.48\textwidth}
         \includegraphics[width=\textwidth]{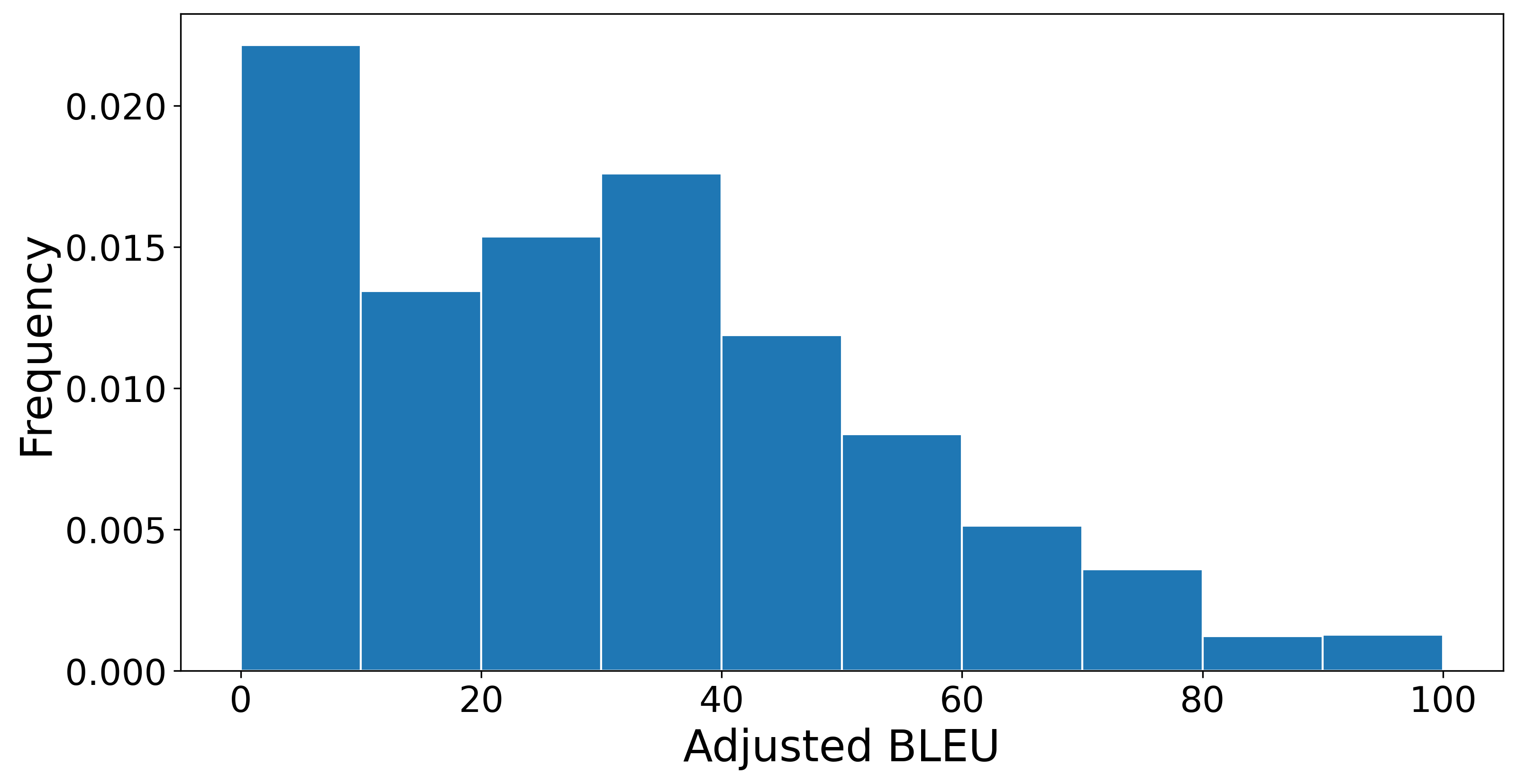}
         \caption{German--English}
     \end{subfigure}
\caption{Adjusted BLEUs for the test data used in the hallucination analysis of Sec.~\ref{sec:explain}.}
\label{fig:hist-bleu}     
\end{figure*}

We motivate here the choice of thresholds used in the hallucination detection approach discussed in Sec.~\ref{sec:explain}. Figure~\ref{fig:hist-bleu} shows an histogram with the normalized frequencies of the values of the adjusted BLEU with the concatenation of all the test data used (for English--German and German--English) in Fig.~\ref{fig:hallucinations}. It can be seen that more than 20\% of the sentences would be regarded as hallucinations by our identification approach; our empirical observations corroborate this point, which may be explained by the low-resource scenario in which our experiments are run.

Table~\ref{tab:examples-adjusted-bleus} shows some examples of references and generated translations together with the corresponding adjusted BLEU scores. It includes two output sentences which are regarded as hallucinations and one that is not.

The sentence-level smoothing approach used when computing the adjusted BLEU scores was based on the common technique~\citep{chen-smooth} of adding 1 to the matched $n$-gram count and the total $n$-gram count for $n$ ranging from 2 to the maximum order of $n$-grams $N$ ($N$ is usually 4, but it is $2$ in our case). Notice that this implies that if no unigram is matched, the resulting BLEU is $0$. We thus consider that if no single token co-occurs, an hallucination is happening. However, bigram counts are smoothed as we do not want them to excessively affect the score. In fact, instead of adding $1$ to both counts, we add $0.1$. This is in line with weighting the precision ratio for bigrams with a weight (0.2) four times smaller than that of unigrams (0.8).  

\begin{table*}[tb]
\centering
\fontsize{10.2}{12.0}\selectfont
\begin{tabular}{clp{0.65\textwidth}}
\textbf{Adjusted BLEU} & \textbf{Type} & \textbf{Sentence} \\
\hline \hline
\multirow{3}{*}{1.74} & input & artikel 1 der verordnung (eg) nr. 1002/2004 erhält folgende fassung: \\ 
                      & reference & article 1 of regulation (ec) no 1002/2004 shall be amended as follows: \\ 
                      & output & articles one of the figure-to-vis-it-vis-vis-a-vis-vis-vis-a-vis-vis-vis-vis-vis-vis-vis-vis-vis-vis-a-vis-vis-vis-vis-a-vis-vis-vis. \\ \midrule
\multirow{3}{*}{28.89} & input & artikel 1 der verordnung (eg) nr. 1002/2004 erhält folgende fassung: \\
                       & reference & article 1 of regulation (ec) no 1002/2004 shall be amended as follows: \\ 
                       & output & articles 1 the requirement (eg) number 1002 / 2004 receives the following framework. \\ \midrule
\multirow{3}{*}{0} & input & verfalldatum \\
                   & reference & expiry date \\ 
                   & output & dr: why don't you think about this? \\                      
\hline
\end{tabular}
\caption{An output sentence, emitted by the baseline system, which was labelled as an hallucination (adjusted BLEU of $1.74$). Below that, a non-hallucinated translation (adjusted BLEU of $28.89$) generated by one of the DA systems. At the bottom, a more extreme example of hallucination with an adjusted BLEU of 0.}
\label{tab:examples-adjusted-bleus}
\end{table*}

\end{document}